\documentclass[letterpaper, 10 pt, conference]{ieeeconf}  

\IEEEoverridecommandlockouts                              

\usepackage{xcolor}
\usepackage{times}
\usepackage{epsfig}
\usepackage{amsmath}
\usepackage{graphicx}
\usepackage{amssymb}
\usepackage{colortbl}
\usepackage{multirow}
\usepackage{soul}
\usepackage{adjustbox}
\usepackage{array}
\usepackage{setspace}
\usepackage{lscape}
\usepackage{mathtools}
\usepackage{hyperref}
\usepackage{pgfplots}
\pgfplotsset{compat=newest}
\let\labelindent\relax
\usepackage{enumitem}
\usepackage{rotating}
\usepackage{caption}
\usepackage{cite}
\usepackage{amsmath,amssymb,amsfonts}
\usepackage{algorithmic}
\usepackage{subcaption}
\usepackage{xcolor}
\usepackage{stix}
\usepackage{cleveref}
\usepackage{comment}
\usepackage{booktabs}
\usepackage{titlesec}
\titlespacing*{\subsubsection}
  {0pt}      
  {1.5em}    
  {0.1em}   

\begin{document}

\title{\LARGE \bf
Exploring Deep Learning and Ultra-Widefield Imaging \\for Diabetic Retinopathy and Macular Edema}





\author{Pablo Jimenez-Lizcano$^{1,\dagger}$, Sergio Romero-Tapiador$^{1,\dagger,*}$, Ruben Tolosana$^{1}$, Aythami Morales$^{1,2}$, \\Guillermo González de Rivera$^{3}$, Ruben Vera-Rodriguez$^{1}$ and Julian Fierrez$^{1}$\\
{\footnotesize $^{1}$BiometricsAI, Universidad Autónoma de Madrid,  Madrid, Spain}\\{\footnotesize $^{2}$Department of Mathematics, Universidad de Las Palmas de Gran Canaria, Spain}\\{\footnotesize $^{3}$HCTLab Research Group, Universidad Autónoma de Madrid, Madrid, Spain}\\{\footnotesize$^{\dagger}$These authors contributed equally to this work. *Corresponding author: \href{mailto:sergio.romero@uam.es}{sergio.romero@uam.es}}}

\maketitle

\thispagestyle{empty}
\pagestyle{empty}

\begin{abstract}

Diabetic retinopathy (DR) and diabetic macular edema (DME) are the leading causes of preventable blindness among working-age adults. Traditional approaches in the literature focus on standard color fundus photography (CFP) for the detection of these conditions. Nevertheless, recent ultra-widefield imaging (UWF) offers a significantly wider field of view compared to CFP. Motivated by this, the present study explores state-of-the-art deep learning (DL) methods and UWF imaging on three clinically relevant tasks: \textit{i)} image quality assessment for UWF, \textit{ii)} identification of referable diabetic retinopathy (RDR), and \textit{iii)} identification of DME. Using the publicly available UWF4DR Challenge dataset, released as part of the MICCAI 2024 conference, we benchmark DL models in the spatial (RGB) and frequency domains, including popular convolutional neural networks (CNNs) as well as recent vision transformers (ViTs) and foundation models. In addition, we explore a final feature-level fusion to increase robustness. Finally, we also analyze the decisions of the DL models using Grad-CAM, increasing the explainability. Our proposal achieves consistently strong performance across all architectures, underscoring the competitiveness of emerging ViTs and foundation models and the promise of feature-level fusion and frequency-domain representations for UWF analysis.
\end{abstract}


\section{Introduction}


Diabetic retinopathy (DR) is a common microvascular complication of diabetes mellitus and the leading cause of preventable blindness in working-age adults\cite{Cheung2010}. In 2020, the number of adults worldwide with DR was estimated at 103.1 million, while by 2045 this figure is projected to increase to 160.5 million\cite{Teo2021}. Among DR complications, diabetic macular edema (DME) is a severe condition characterized by accumulation of fluid and hard exudates within the macula, resulting from microvascular leakage and leading to retinal thickening and central vision impairment \cite{Elyasi2021}. Early detection is therefore crucial to reduce their burden. In this context, artificial intelligence (AI) has shown remarkable potential for automated retinal analysis. However, while standard color fundus photography (CFP) is the reference for screening, its limited field of view (30$^{\circ}$--50$^{\circ}$) restricts the evaluation of peripheral lesions~\cite{ETDRS1991}. In contrast, ultra-widefield (UWF) imaging captures up to 200$^{\circ}$ of the retina in a single image, providing a broader context for clinical evaluation \cite{Sun2016}. Despite this advantage, deep learning (DL) on UWF remains limited compared to the extensive literature on CFP.

\begin{figure*}[!t]
    \centering
    \includegraphics[width=0.89\textwidth]{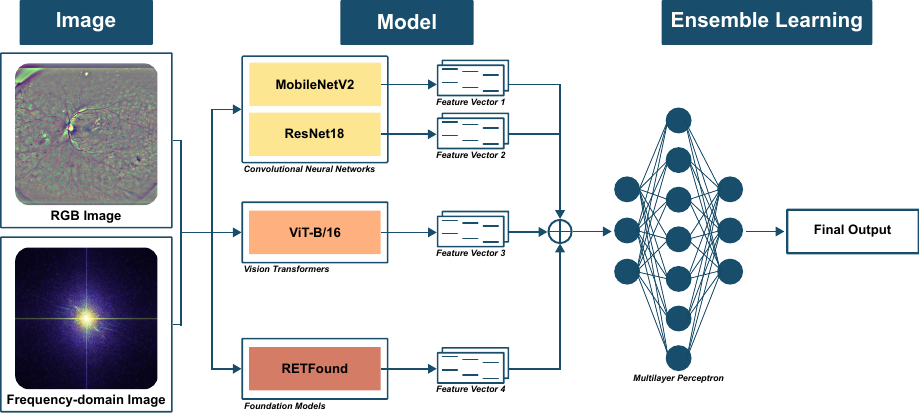}
\caption{Graphical representation of the proposed framework, summarizing the main components and processing stages of the approach. Both RGB and frequency-domain images are considered in the analysis. Four state-of-the-art deep learning models are studied (MobileNetV2, ResNet18, ViT-B/16, and RETFound). Feature vectors are then concatenated to form a unified multimodal embedding, which is finally fed into a multilayer perceptron (MLP).}
\label{fig:ensemble_fusion}
\end{figure*}

The present study addresses this gap by conducting a systematic study of state-of-the-art DL methods using ultra-widefield fundus imaging for diabetic retinopathy (UWF4DR) Challenge dataset. We focus on three clinically relevant tasks: \textit{i)} image quality assessment for UWF, \textit{ii)} identification of referable diabetic retinopathy (RDR), and \textit{iii)} identification of DME. Although most previous studies have focused on CFP and relied primarily on convolutional neural networks (CNNs) trained on RGB information, we expanded the state of the art by exploring spatial- and frequency-domain representations using CNNs, vision transformers (ViTs), and foundation models.  In addition, we explore feature-level fusion to improve robustness and use Grad-CAM~\cite{GradCAM2017} to increase explainability. Fig.~\ref{fig:ensemble_fusion} provides a graphical representation of the proposed framework, summarizing the main components and processing stages of the approach. It is important to note that all all details regarding the experimental protocol and the proposed DL models are available on GitHub\footnote{\url{https://github.com/BiDAlab/UWF4DR-Benchmark}} for reproducibility reasons.


\begin{table*}[t]
\centering
\caption{Experimental protocol: Distribution of images across training, validation, and test subsets of UWF4DR dataset.}
\begin{tabular}{l *{9}{c}}
\toprule
& \multicolumn{3}{c}{Task 1: Quality Assessment} & \multicolumn{3}{c}{Task 2: RDR Identification} & \multicolumn{3}{c}{Task 3: DME Identification} \\
\cmidrule(lr){2-4} \cmidrule(lr){5-7} \cmidrule(lr){8-10}
& Ungradable & Gradable & Total & No RDR & RDR & Total & No DME & DME & Total \\
\midrule
Training & 138 & 179 & 317 & 69 & 92 & 161 & 75 & 61 & 136 \\
Validation & 42 & 37 & 79 & 16 & 24 & 40 & 15 & 19 & 34 \\
Test & 49 & 50 & 99 & 25 & 25 & 50 & 21 & 21 & 42 \\
\bottomrule
\end{tabular}
\label{tab:dataset_summary}
\end{table*}


The remainder of this paper is organized as follows. Section \ref{related_work} reviews related work on deep learning for retinal imaging. Section \ref{material_methods} introduces the UWF4DR Challenge dataset and the experimental protocol. Section \ref{proposed_method} details the proposed methodology. Section \ref{results_discussion} presents and discusses the results. Finally, Section \ref{conclusion_future_work} concludes the paper and outlines future directions.

\section{Related Work}\label{related_work}

DL has been applied on CFP for the identification of DR and DME. The seminal works by Gulshan \textit{et al.}\cite{Gulshan2016}, Ting \textit{et al.}\cite{Ting2017}, and Gargeya and Leng\cite{Gargeya2017} achieved ophthalmologist-level performance using CNN ensembles, establishing CFP as the reference standard despite its limited field of view.

Compared with CFP, research on UWF imaging remains limited, but has explored several complementary directions. Regarding disease classification, studies have employed architectures like DenseNet and EfficientNet to detect vascular diseases with high accuracy \cite{Abitbol2022,Sun2022}, while Oh \textit{et al.} \cite{Oh2021} incorporated segmentation to improve detection. In terms of image quality, Calderon-Auza \textit{et al.} \cite{Calderon2020} and Li \textit{et al.}~\cite{Li2020} developed CNN pipelines to identify artifacts and assess structure visibility. In parallel, emerging architectures such as ViTs \cite{Nazih2023,Yang2024} and foundation models \cite{Zhou2023} have shown superior generalization capabilities in medical imaging and, in particular, in retinal analysis. Despite these advances, their application to UWF imaging remains underexplored.

\section{UWF4DR Challenge Dataset}\label{material_methods}

The present study considers the publicly available UWF4DR Challenge dataset\footnote{\scriptsize \url{https://codalab.lisn.upsaclay.fr/competitions/18605}}, released as part of the MICCAI 2024 conference. The competition aimed to benchmark DL methods on UWF images acquired with Optos devices for clinically relevant tasks in DR and DME\footnote{\scriptsize \url{https://zenodo.org/records/10992021}}. The UWF4DR Challenge is structured into three binary tasks that reflect key stages and clinically relevant decision points in a realistic UWF-based automated screening workflow: 

\begin{itemize}
    \item \textbf{Task 1 - Quality Assessment:} Classification of images as \textit{Gradable} (clear visualization of optic disk/macula) or \textit{Ungradable} (artifacts, blur, or severe occlusion).
    \item \textbf{Task 2 - RDR Identification:} Distinguishes between \textit{Non-referable} (healthy/mild DR) and \textit{Referable DR} (moderate DR or worse/DME).
    \item \textbf{Task 3 - DME Identification:} Identifies the presence of DME (retinal thickening/exudates near the fovea).
\end{itemize}

\begin{figure*}[t]
    \centering
    \includegraphics[width=0.8\textwidth]{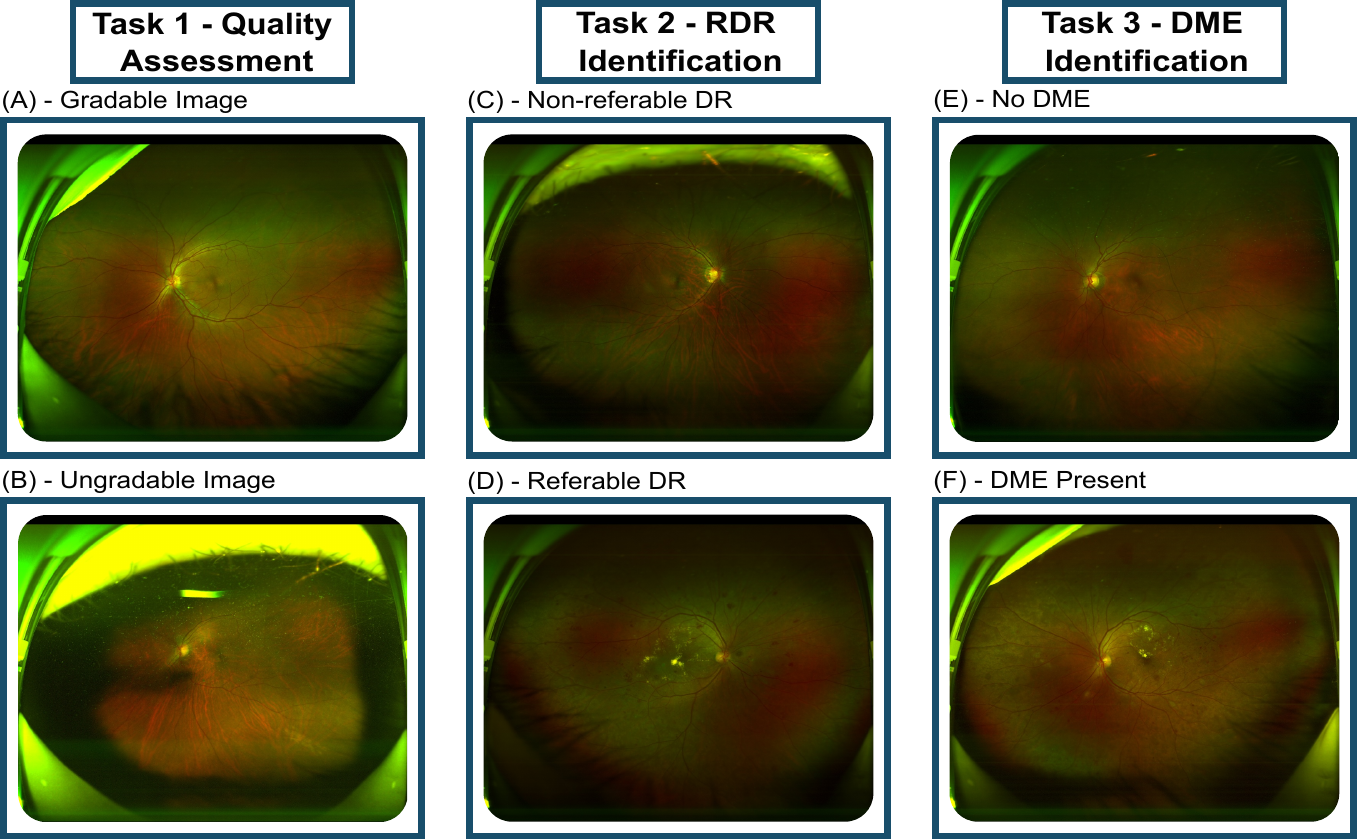}
    \captionsetup{skip=22pt}
    \caption{
    UWF4DR tasks.
    (A--B) Quality Assessment: (A) Gradable image with good focus, contrast, and clear structures; (B) Ungradable due to blur, media opacities, and partial eyelid obstruction.
    (C--D) RDR Identification: (C) Non-referable case; (D) RDR case exhibiting intraretinal hemorrhages, hard exudates, and microaneurysms.
    (E--F) DME Identification: (E) No DME (normal macula); (F) DME present, showing hard exudates and macular blurring due to fluid accumulation.
    }
    \label{fig:tasks_examples}
\end{figure*}

Visual examples are shown in Fig.~\ref{fig:tasks_examples}. Since the official test set for the UWF4DR Challenge is withheld, and by this work was performed, the official competition and its rankings had already been finalized, we decided to establish an independent experimental protocol, randomly splitting the data into subsets of training (64\%), validation (16\%) and test (20\%) (see Table~\ref{tab:dataset_summary}). Specific details are available in our GitHub repository for reproducibility reasons.

\begin{figure}[t]
    \centering
    \includegraphics[width=0.8\linewidth]{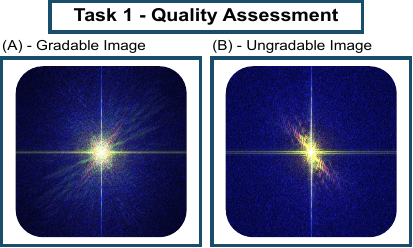}
    \vspace{1pt}
    \caption{DFT Magnitude (clipped 99\%) comparison: (A) Gradable image showing balanced frequencies; (B) Ungradable image showing concentration in low frequencies due to blur.}
    \label{fig:dft_comparison}
\end{figure}

\section{Proposed Method}\label{proposed_method}

Fig.~\ref{fig:ensemble_fusion} provides a graphical representation of the proposed method. Unlike previous approaches in the literature, in the present study we explore two input image domains:

\begin{itemize}
    \item \textbf{Spatial Domain (RGB):} Images are cropped to $800 \times 800$ to remove background while preserving peripheral view ($448 \times 448$ for Task 1). Local mean subtraction \cite{Zhang2024} is applied for color normalization. Data augmentation (flips, rotations, zoom) mitigates overfitting.
    \item \textbf{Frequency Domain:} To capture texture anomalies (e.g., blur in Task 1), we compute the 2D DFT magnitude clipped at the 99th percentile to suppress noise. As illustrated in Fig.~\ref{fig:dft_comparison}, gradable images exhibit a balanced distribution with predominant mid-frequencies (indicating sharpness), while ungradable inputs concentrate energy in low frequencies with greater high-frequency dispersion (reflecting blur and noise).
\end{itemize}

In addition, we explore a diverse set of DL architectures in UWF images, covering CNNs, transformers, and foundation models. Specifically, MobileNetV2\cite{Sandler2018} and ResNet18\cite{He2016} are selected as representative CNNs; ViT-B/16\cite{Dosovitskiy2020} is a ViT that leverages self-attention mechanisms to capture long-range dependencies; and RETFound\cite{Zhou2023} is a large-scale foundation model pre-trained on millions of retinal images with self-supervised learning and masked autoencoders (MAE). This combination of architectures is chosen to assess whether emerging transformer-based and foundation approaches achieve performance comparable to state-of-the-art CNNs and to explore their suitability for UWF imaging analysis in terms of classification performance and clinical relevance. All experiments are implemented in TensorFlow/Keras. A two-stage fine-tuning strategy is adopted for CNNs and ViTs: in the first stage, only the newly added dense layers are trained, with the backbone (pre-trained on ImageNet database \cite{deng2009imagenet}) frozen; in the second stage, selected deeper layers are unfrozen and jointly optimized. RETFound is adapted by replacing its classification head with a task-specific multilayer perceptron (MLP), training the encoder and MLP with CutMix augmentation. We use the AdamW optimizer (learning rate and weight decay of $10^{-4}$), categorical cross-entropy loss, and early stopping based on AUROC (validation subset). DL models publicly available\footnote{\url{https://github.com/BiDAlab/UWF4DR-Benchmark}}.

Finally, to further enhance robustness, we explore ensemble learning strategies based on feature-level fusion of the models trained independently on RGB and on frequency-domain representations, which allows leveraging the complementary representations learned by different DL architectures. Specifically, feature vectors from all selected models are extracted from intermediate layers, standardized, and concatenated in a single vector. An MLP is then trained on top of these combined features to generate the final output. This strategy, illustrated in Fig.~\ref{fig:ensemble_fusion}, is applied separately to both the spatial and frequency domains.

\section{Results and Discussion}\label{results_discussion}

This section evaluates our proposed approach on the test set using popular metrics in the literature such as AUROC, AUPRC, sensitivity, and specificity. Table~\ref{tab:all_tasks_results} summarizes the performance in all three tasks and architectures.

\subsection{Task 1: Image Quality Assessment}
The RGB-based models achieve strong performance, with AUROC values greater than 93.0\%. Among them, ViT-B/16 achieves the highest AUROC (95.4\%), closely followed by ResNet18 (94.8\%). The best overall performance is obtained by the RGB ensemble, which achieves the top AUROC (96.4\%), AUPRC (97.1\%), and specificity (98.0\%). In contrast, frequency-domain models provide consistently lower performance, with values between 80.0\% and 86.0\% AUROC, although the frequency-domain ensemble improves over individual models, reaching 87.8\% AUROC and specificity. This high specificity is clinically critical to reliably exclude contaminable scans, thus reducing the risk of missed lesions due to poor visibility. Furthermore, frequency-domain models demonstrate that spectral features retain discriminative power to detect blur and noise.

\begin{figure*}[t]
\centering
\includegraphics[width=0.75\textwidth]{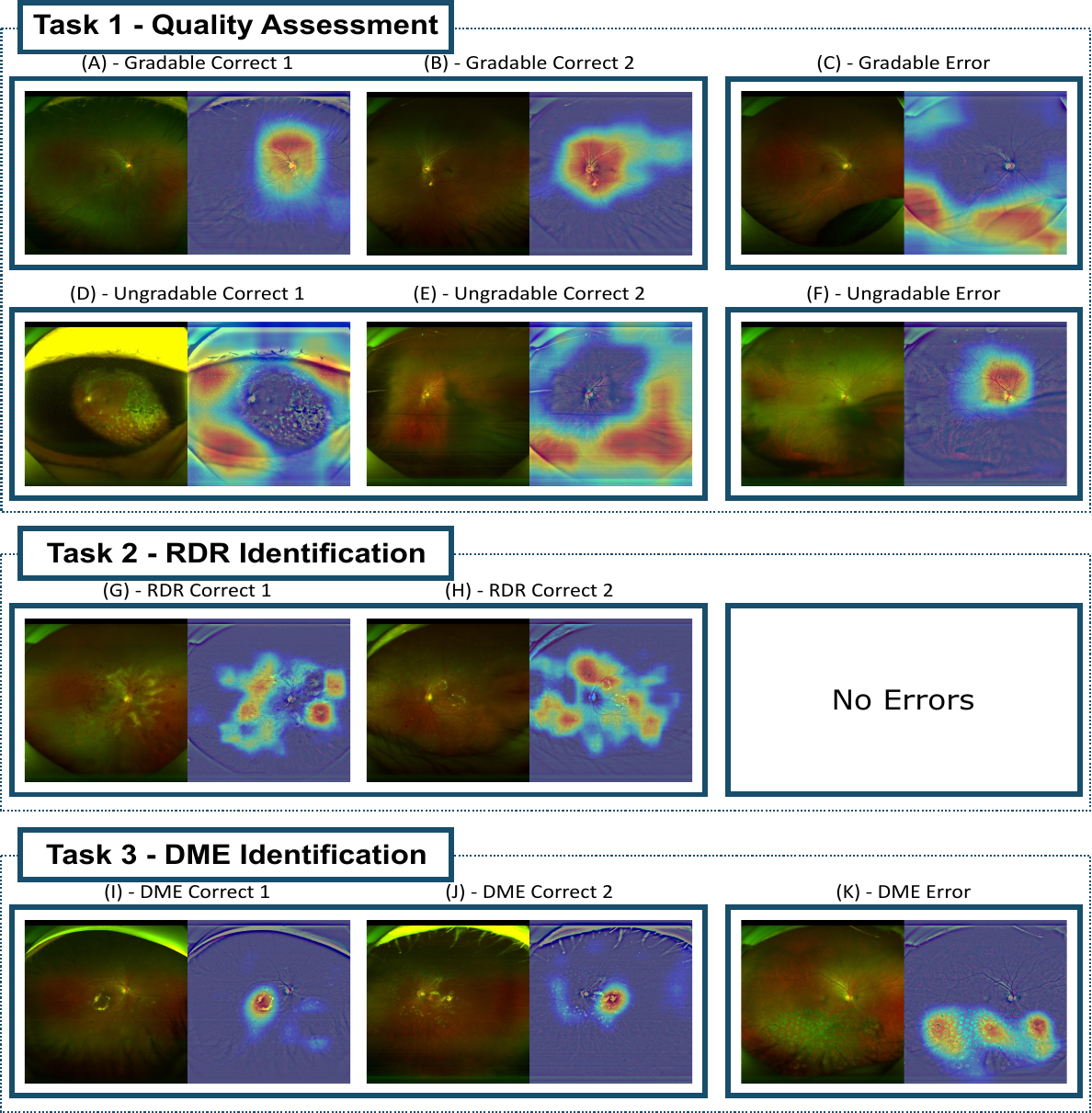}
\captionsetup{skip=20pt}
\caption{Grad-CAM examples. Task 1 (Quality Assessment): (A--B) Correct gradable predictions focus on optic disc/vessels; (D--E) Correct ungradable predictions highlight peripheral opacities/eyelids; (C, F) Misclassifications. Task 2 (RDR Identification): (G--H) Attention on characteristic lesions (hemorrhages, exudates). Task 3 (DME Identification): (I--J) Focus concentrated on the macular region; (K) Misclassification. Each panel displays the  UWF image and its Grad-CAM heatmap.}
\label{fig:gradcam_examples}
\end{figure*}

In order to provide a better explanation to the doctors, we analyze the decisions of the DL models considering Grad-CAM\cite{GradCAM2017} visualization techniques. This method highlights the image regions that contribute the most to the decision of the model, allowing us to assess whether the predictions are based on clinically significant retinal features. The resulting Grad-CAM maps show that the proposed DL models focus on the optic disk and vascular arcades when classifying gradable images (Fig.~\ref{fig:gradcam_examples}A–B), in agreement with clinical practice\cite{Li2020}. Similarly, for ungradable correct cases, the proposed DL models focus on darkened or blurred regions, which are clinically relevant indicators of poor quality (Fig.~\ref{fig:gradcam_examples}D–E). In contrast, misclassified examples reflect less reliable attention: in Fig.~\ref{fig:gradcam_examples}C, the model emphasizes the eyelid and eyelashes occluding part of the field of view, while in Fig.~\ref{fig:gradcam_examples}F it overlooks a central opacity, leading to misclassification.

\begin{table}[t!]
\centering
\caption{Performance of all DL models across the three tasks and two input image domains. Results are reported in terms of AUROC, AUPRC, Sensitivity (Sens.), and Specificity (Spec.). 
The best metric values for each task and domain are highlighted in \textbf{bold}.}
\label{tab:all_tasks_results}
\begin{tabular}{lcccc}
\toprule
\textbf{DL Model} & \textbf{AUROC} & \textbf{AUPRC} & \textbf{Sens.} & \textbf{Spec.} \\
\addlinespace[0.5em]
\midrule
\multicolumn{5}{c}{\textbf{Task 1 - Quality Assessment: RGB domain}} \\ 
\cmidrule(lr){1-5}
MobileNetV2 & 94.4\% & 95.4\% & \textbf{90.0\%} & 91.8\% \\
ResNet18    & 94.8\% & 95.3\% & 88.0\% & 91.8\% \\
ViT-B/16    & 95.4\% & 96.0\% & 88.0\% & 91.8\% \\
RETFound    & 93.3\% & 94.0\% & \textbf{90.0\%} & 93.9\% \\
Fusion (RGB) & \textbf{96.4\%} & \textbf{97.1\%} & 86.0\% & \textbf{98.0\%} \\
\midrule
\multicolumn{5}{c}{\textbf{Task 1 - Quality Assessment: Frequency domain}} \\ 
\cmidrule(lr){1-5}
MobileNetV2 & 86.3\% & \textbf{88.8\%} & \textbf{86.0\%} & 73.5\% \\
ResNet18    & 84.1\% & 80.8\% & 72.0\% & 83.7\% \\
ViT-B/16    & 83.6\% & 81.3\% & \textbf{86.0\%} & 73.5\% \\
RETFound    & 80.4\% & 83.1\% & 72.0\% & 81.6\% \\
Fusion (Freq.) & \textbf{87.8\%} & 87.0\% & 80.0\% & \textbf{87.8\%} \\
\addlinespace[1.2em]
\midrule
\multicolumn{5}{c}{\textbf{Task 2 – RDR Identification: RGB domain}} \\ 
\cmidrule(lr){1-5}
MobileNetV2 & 99.0\% & 99.2\% & 96.0\% & \textbf{100\%} \\
ResNet18    & 99.4\% & 99.5\% & 96.0\% & \textbf{100\%} \\
ViT-B/16    & 99.5\% & 99.6\% & 96.0\% & \textbf{100\%} \\
RETFound    & 99.5\% & 99.5\% & \textbf{100\%} & 96.0\% \\
Fusion (RGB) & \textbf{100\%} & \textbf{100\%} & \textbf{100\%} & \textbf{100\%} \\
\midrule
\multicolumn{5}{c}{\textbf{Task 2 – RDR Identification: Frequency domain}} \\ 
\cmidrule(lr){1-5}
MobileNetV2 & 92.5\% & 94.4\% & \textbf{88.0\%} & 92.0\% \\
ResNet18    & 92.0\% & 94.5\% & 80.0\% & 96.0\% \\
ViT-B/16    & 90.9\% & 93.2\% & 84.0\% & 92.0\% \\
RETFound    & 90.9\% & 91.9\% & 84.0\% & 92.0\% \\
Fusion (Freq.) & \textbf{92.6\%} & \textbf{94.7\%} & 76.0\% & \textbf{100\%} \\
\addlinespace[1.2em]
\midrule
\multicolumn{5}{c}{\textbf{Task 3 – DME Identification: RGB domain}} \\ 
\cmidrule(lr){1-5}
MobileNetV2 & 92.5\% & 92.9\% & 90.5\% & 85.7\% \\
ResNet18    & 96.6\% & 96.2\% & \textbf{100\%} & 85.7\% \\
ViT-B/16    & 93.0\% & 95.0\% & 85.7\% & \textbf{95.2\%} \\
RETFound    & 95.9\% & 95.9\% & 95.2\% & 85.7\% \\
Fusion (RGB) & \textbf{96.8\%} & \textbf{96.9\%} & 90.5\% & \textbf{95.2\%} \\
\midrule
\multicolumn{5}{c}{\textbf{Task 3 – DME Identification: Frequency domain}} \\ 
\cmidrule(lr){1-5}
MobileNetV2 & 85.9\% & 87.8\% & 76.2\% & 81.0\% \\
ResNet18    & 83.9\% & 89.5\% & 81.0\% & 85.7\% \\
ViT-B/16    & 80.1\% & 81.4\% & \textbf{90.5\%} & 61.9\% \\
RETFound    & 84.4\% & 80.4\% & 71.4\% & 95.2\% \\
Fusion (Freq.) & \textbf{89.3\%} & \textbf{92.9\%} & 76.2\% & \textbf{100\%} \\
\bottomrule
\end{tabular}
\end{table}

\subsection{Task 2: RDR Identification}

As can be seen in Table~\ref{tab:all_tasks_results}, RGB models achieve near-perfect performance (AUROC $>$ 99.0\%), with the ensemble achieving 100\% in all metrics. By contrast, frequency-domain models perform less poorly, with AUROCs ranging from 90.9\% to 92.5\%, although the ensemble improves this to 92.6\% with perfect specificity.

In Task 2 we can observe the best results across all domains. High sensitivity is crucial for screening programs to avoid false negatives. Performance reflects that referable DR lesions, such as hemorrhages and neovascularization, are clearly captured in UWF. Although frequency models achieve lower scores, they provide robust detection, confirming that Fourier features offer complementary information. The small gap between CNNs and emerging models indicates that both families effectively capture diagnostic lesions.

Grad-CAM (Fig.~\ref{fig:gradcam_examples}G–H) supports these results, as correct predictions highlight retinal areas containing hemorrhages and exudates, matching clinical markers\cite{Wilkinson2003}. This confirms that the models detect characteristic UWF lesions and rely on clinically significant features.

\subsection{Task 3: DME Identification}
As can be seen in Table~\ref{tab:all_tasks_results}, the DME identification task seems to be the most challenging task. Among RGB models, ResNet18 maximizes sensitivity (100\%) and ViT-B/16 specificity (95.2\%), while the ensemble achieves the best overall results (AUROC 96.8\%, AUPRC 96.9\%). Similarly to previous tasks, the frequency models perform lower.

RGB models show varying sensitivity-specificity trade-offs, reflecting a subtler occurrence of macular exudates in UWF compared to CFP. Frequency-based models struggle as Fourier transforms dilute local details, but ensembles enhance robustness. Small performance gaps across CNNs, ViTs, and RETFound suggest that all of them are adequate.

Grad-CAM  (Fig. \ref{fig:gradcam_examples}I–J) highlights macular exudates, aligning with clinical definitions \cite{Elyasi2021}. In errors (Fig.~\ref{fig:gradcam_examples}K),  attention drifts to confounding greenish features (e.g., scars or artifacts), explaining reduced sensitivity.

\subsection{Discussion}
Although RGB representations remain the most reliable for UWF analysis, frequency provides complementary cues that can enhance models, especially through ensemble strategies. The absence of clear differences between conventional CNNs and emerging architectures suggests that multiple model families can be effectively applied in this setting, which encourages the deployment of DL systems in clinical workflows. In addition, the Grad-CAM visualizations further support our findings, showing that the predictions relied on clinically meaningful retinal structures, reinforcing the potential integration of these models into ophthalmic workflows.

Despite our contributions, some limitations still remain. First, experiments are based on a single database, limiting their generalizability. Furthermore, the organizers of the UWF4DR Challenge formulated the three tasks as binary classifications (e.g., presence vs. absence of RDR or DME), whereas real-world screening requires severity grading.

\section{Conclusion and Future work}\label{conclusion_future_work}

In the present study, we have explored state-of-the-art DL models and recent UWF imaging for three clinically relevant tasks: image quality assessment, RDR identification, and DME identification, using the publicly available UWF4DR Challenge dataset. We have benchmarked both conventional CNNs and emerging architectures, such as ViTs and foundation models, while also exploring frequency-domain representations and feature-level fusion. Although DL models trained on RGB images consistently have outperformed those using frequency information, the latter still has achieved satisfactory results and showed promise as a complementary representation. Overall, all evaluated architectures have delivered strong performance, with fusion strategies often improving robustness. Importantly, explainability based on Grad-CAM visualization techniques has revealed that the proposed DL models' decisions are largely aligned with clinically relevant retinal structures, supporting their potential integration into ophthalmic workflows.

Future work should consider:  \textit{i)} extending evaluations to additional UWF datasets as they become available, including synthetic data \cite{melzi2023synthetic, romero2023ai4food,  shahreza2024sdfr}, \textit{ii)} exploring multi-class classification of DR severity, \textit{iii)} increasing performance and explainability \cite{deandres2024pixels} through novel vision-language models (VLMs) \cite{romero2025vision, deandres2024good} and methods to extract better visual features \cite{dealcala2025attzoom}, and \textit{iv)} advancing fusion strategies to better combine RGB and frequency-domain information. Beyond technical aspects, prospective clinical validation will be critical to assess the utility, safety, and fairness of these systems. Our future work will also explore how to properly curate \cite{pena2024continuous} our AI-based methods using human inputs \cite{2023_SNCS_Human-Centric_Pena}.

\section*{Acknowledgments}
Support by PowerAI+ (SI4/PJI/2024-00062 Comunidad de Madrid and UAM), Cátedra ENIA UAM-Veridas en IA Responsable (NextGenerationEU PRTR TSI-100927-2023-2), and TRUST-ID (PID2025-173396OB-I00 MICIU/AEI and the EU). In addition, we acknowledge the computer resources provided by the Centro de Computación Cient ífica-UAM. Research conducted in the ELLIS Unit Madrid.

{
\bibliographystyle{IEEEtran}
\bibliography{references/refs}
}


\end{document}